%% file: aaai25.tex
\documentclass[letterpaper]{article} 
\usepackage{aaai25}  
\usepackage{times}  
\usepackage{helvet}  
\usepackage{courier}  
\usepackage[hyphens]{url}  
\usepackage{graphicx} 
\usepackage{natbib}  
\usepackage{caption} 
\usepackage{algorithm}
\usepackage{algorithmic}
\usepackage{amsmath}
\usepackage{booktabs}  
\usepackage{amssymb}

\usepackage{newfloat}
\usepackage{listings}
\DeclareCaptionStyle{ruled}{labelfont=normalfont,labelsep=colon,strut=off} 
\floatstyle{ruled}
\newfloat{listing}{tb}{lst}{}
\floatname{listing}{Listing}
\title{NeuRN: Neuro-inspired Domain Generalization\\for Image Classification}
\author {
    Hamd Jalil\thanks{Equal contribution.}\quad
    Ahmed Qazi\footnotemark[1]\quad
    Asim Iqbal\thanks{Corresponding author.}
}
\affiliations {
    \normalfont
    \Large
    Tibbling Technologies\\
    \tt\small asim@tibbtech.com
}

\begin{document}

\maketitle

\begin{abstract}
Domain generalization in image classification is a crucial challenge, with models often failing to generalize well across unseen datasets. We address this issue by introducing a neuro-inspired \textbf{Neu}ral \textbf{R}esponse \textbf{N}ormalization (\textbf{NeuRN}) layer which draws inspiration from neurons in the mammalian visual cortex, which aims to enhance the performance of deep learning architectures on unseen target domains by training deep learning models on a source domain. The performance of these models is considered as a baseline and then compared against models integrated with NeuRN on image classification tasks. We perform experiments across a range of deep learning architectures, including ones derived from Neural Architecture Search and Vision Transformer. Additionally, in order to shortlist models for our experiment from amongst the vast range of deep neural networks available which have shown promising results, we also propose a novel method that uses the Needleman-Wunsch algorithm to compute similarity between deep learning architectures. Our results demonstrate the effectiveness of NeuRN by showing improvement against baseline in cross-domain image classification tasks. Our framework attempts to establish a foundation for future neuro-inspired deep learning models.
\end{abstract}

\section{Introduction}

Recent advances in deep learning have revolutionized the field of machine learning, leading to unprecedented breakthroughs across a range of applications. These advances, particularly in neural network architectures and training methodologies, have resulted in state-of-the-art performance in various real-world applications, notably in computer vision \cite{11, 13, 14, 15, 16}. Deep learning's success is largely attributed to its ability to learn complex, high-dimensional representations from large datasets and achieve remarkable accuracy in tasks such as image recognition, object detection, and semantic segmentation. However, a critical challenge that persists in the field is the problem of domain generalization which refers to the ability of a model trained on a specific (source) dataset (or domain) to perform well on unseen (target) datasets \cite{blanchard2021domain, taori2020measuring,
ben-david2010theory, recht2019imagenet, moreno-torres2012unifying}. This is largely because the real world presents an enormous variety of data, often with varying distributions as compared to the training data. Consequently, deep learning models often suffer from a drop in performance when confronted with new domains in test, an effect known as domain shift.


The pursuit to overcome the problem of domain shift in deep learning has led to the development of architectures and techniques aimed at enhancing domain generalization abilities of Deep Neural Networks (DNNs) \cite{recht2019imagenet, wang2022generalizing} as well as testing their biological significance \cite{iqbal2019decoding}. Despite these efforts, many architectures are still tailored to perform well within specific domains and often lack transferability to unseen domains. Neuro-inspired models have been used to solve this problem, e.g. CNNs have been shown to be able to predict neural responses in the mammalian visual cortex (MVC) \cite{Cadena2019HowWD}. Studies show CNN models biased towards a more biological feature space lead to increased robustness \cite{50}. Thus, expanding upon MVC principles and drawing from the Winner-Takes-All (WTA) mechanism \cite{Iqbal2024}, we introduce a novel, brain-inspired approach for domain generalization. Our technique, \textbf{NeuRN} (\textbf{Neu}ral \textbf{R}esponse \textbf{N}ormalization) generates a domain-agnostic feature representation of images by normalizing pixel-level-based contrastive deviations. The Methods section explains the mathematical foundation of our proposed technique. To test its effectiveness, we incorporate NeuRN as a pre-processing layer across a range of DNNs, including Convolutional Neural Networks (CNNs) \cite{goodfellow2016deep}, Vision Transformer (ViT) (\cite{33}) and Neural Architecture Search (NAS) \cite{29} derived models, offering a generalizable solution to the domain shift problem. 

\begin{figure*}
    \centering
    \includegraphics[scale=0.80]{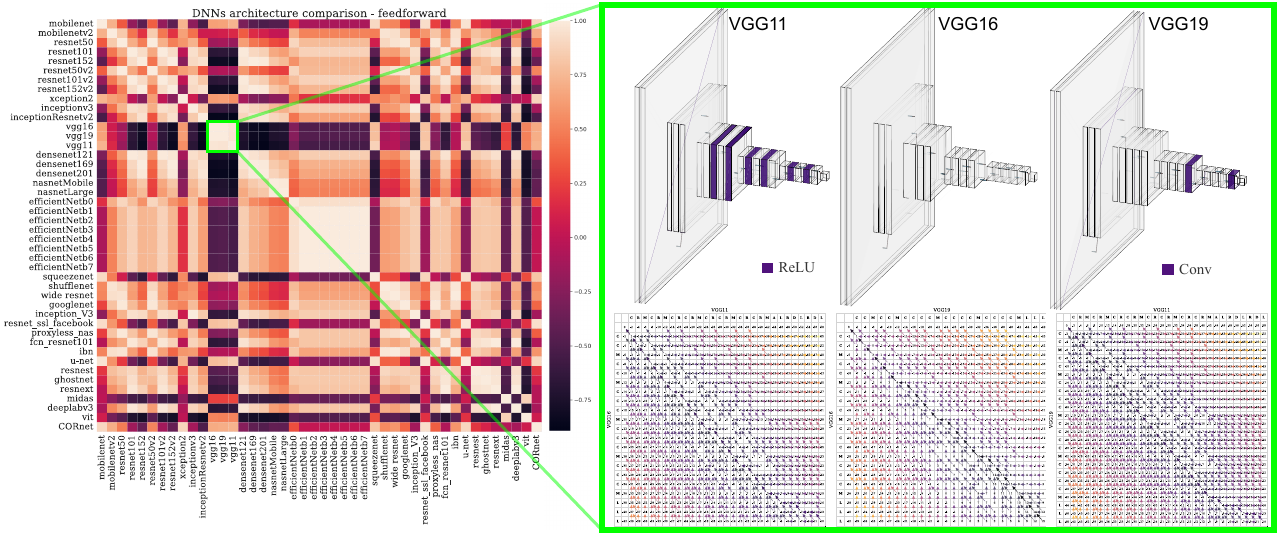}
    \caption{DNNs' architectural similarity: The confusion matrix on the left shows the correlation similarity of 44 DNN architectures, with darker clusters representing weaker correlation and lighter clusters representing stronger correlation. The panel on the right shows what the cells in the matrix represent: a 1:1 architectural comparison between artificial neural networks. The degree of similarity is calculated using the N-W algorithm, and every single layer is compared to build the similarity score. 
    }
    \label{fig:4}
\end{figure*}

From the vast array of DNNs bench-marked on image classification datasets, we shortlist representative models for our experiments. We employ the Needleman-Wunsch (N-W) algorithm \cite{42} to quantify architectural similarities between DNNs. This method serves two purposes: first, it allows us to select models incorporating architectural features from existing DNNs to evaluate our proposed technique, NeuRN. Second, it enhances understanding of DNN architectures by revealing their architectural similarities, and in turn, paves the way for future innovations in neural network design and neuro-inspired DNN development. Our major contributions are summarized as follows:
\\\\
\noindent\textbf{DNN Architectural Similarity:} We introduce a method using the Needleman-Wunsch algorithm to quantify similarity among deep learning models based on the types of layers and recurring sections of layers. This allows us to identify a group of similar architectures and select one representative DNN from each group, enabling efficient experimentation which incorporates the maximum architectural features of existing DNNs.
\\\\
\noindent\textbf{NeuRN}: Drawing parallels with computational mechanisms observed in the visual neurons of the MVC \cite{Iqbal2024}, we propose NeuRN, a novel neuro-inspired layer that can be integrated into any deep learning model processing visual inputs. NeuRN demonstrates broad applicability across DNN solutions and improves performance on image classification tasks where DNNs are trained on a “source-only” domain and tested on “unseen” target domains.

\section{Related Work}
\noindent\textbf{Domain Generalization:} Researchers have explored strategies beyond specialized architectures to enhance the generalizability of deep learning models, including data augmentation \cite{nazari2020domain}, meta-learning \cite{ chen2022discriminative, wang2021meta, zhao2021learningCVPR}, invariant feature learning \cite{segu2020batch, jin2021style, mahajan2021domain, zhangRobust}, and domain alignment techniques \cite{muandet2013domain, motiian2017unified, wang2020respecting, li2020domain}. Data augmentation modifies training data to simulate variations, enhancing robustness; meta-learning enables rapid adaptation with minimal data; invariant feature learning extracts stable features across domains; and domain alignment aligns feature distributions to minimize discrepancies \cite{wang2022generalizing, Zhou_2022}. However, these strategies may focus on superficial variations, missing deeper intrinsic data structures; data augmentation might introduce unrealistic changes, and domain alignment may overlook fine-grained contrasts crucial for generalizability. In contrast, the human visual cortex efficiently generates invariant representations from minimal data through few-shot learning.
\\\\
\noindent\textbf{Deep Learning and Neuroscience:} Integrating deep learning with neuroscience advances the understanding of visual mechanisms and aids the development of artificial neural networks. Comparing DNNs with biological neural responses shows artificial systems emulating biological vision \cite{49}. Studies highlight representational similarities between DNN features and neural activity in response to visual stimuli \cite{21, 22, 26, 38}. Models like CORnet reflect cortical visual processing \cite{8,9}, MouseNet mirrors MVC functionality \cite{45}, and CVSNet implements the brain’s central visual system, showcasing bio-inspired models’ potential \cite{cvsnet2023}. These works unravel neural responses’ relevance to deep learning models. Brain-inspired learning improves DNN performance, helping them mimic the brain’s efficiency and adaptability in visual processing \cite{veerabadranBioInspired}. For instance, \cite{50} used DNNs to probe the mouse visual cortex, enhancing robustness against adversarial attacks. Despite their theoretical significance, these models often struggle in practical applications.
\\\\
We address the need to integrate neuroscientific principles into effective domain generalization models suitable for real-world conditions by introducing NeuRN (Neural Response Normalization). NeuRN leverages biological insights to create a domain-agnostic data representation focused on practical performance. By normalizing image data at the pixel level, it preserves intrinsic structural and contrastive features, enabling robust cross-domain generalization. NeuRN integrates easily into existing DNNs as a pre-processing layer without extensive re-engineering, allowing DNNs to learn domain-agnostic feature representations. The term ``domain-agnostic" here refers to the ability to discover latent domains in the
target domain and exploit the information, making it more broadly applicable and potentially enhancing the DNNs cross-domain generalizability.

\section{Methods}
\subsection{Neural Response Normalization (NeuRN)}
\label{sec:neurn}
Taking inspiration from the Winner-Takes-All (WTA) mechanism observed in neuronal circuits of visual neurons \cite{Iqbal2024}, we introduce Neural Response Normalization (NeuRN), a neuro-inspired layer that encapsulates the response normalization property observed in a specific group of excitatory neurons in the MVC \cite{zayyad2023normalization, burg2021learning, sawada2016emulating, das2021effect}. What distinguishes these neurons is their ability to encode both the structure and contrast of the presented stimuli. This unique encoding is evident from their dynamic spiking profile, which aligns proportionally with the contrasting features of the stimuli. To implement this, we use a high-level mathematical formulation and integrate it as a pre-processing layer in deep learning models.

Consider an input image denoted by \(I \in \mathbb{R}^{W \times H \times C}\), where \(W\), \(H\), and \(C\) represent the width, height, and channel of the image, respectively. We aim to obtain a domain-agnostic representation, \(I_{a}\), of this image. The procedure to derive this representation commences by extracting patches of size \(k\) from each channel in \(I\). Assuming a stride of \(1\), the total number of these patches will equal the pixel count of the input. We can represent the collection of these patches as \(P = \{p_{k_1}, p_{k_2}, \dots, p_{k_n}\}\), where \(p_k\) is defined as the patch of size \(k\) encircling the pixel \(x\) located at the coordinates \(i, j\) within the image. Subsequently, we determine the mean \(\mu_{p_k}\) of each patch as:
\[
\mu_{p_k} = \frac{1}{k^2} \sum_{i,j \in p_k} x_{ij}, \quad \sigma_{p_k} = \sqrt{\frac{1}{k^2} \sum_{i,j \in p,k} (x_{ij} - \mu_{p_k})^2}
\]
The standard deviation \(\sigma_{p_k}\) for the patch located at \(i, j\) is calculated as shown above. Finally, the value of each pixel in the domain-agnostic representation, \(I_a\), of the image is computed using the formula below.
\[
I_a = \frac{1}{c} \cdot \sigma_{p_k}, \text{ where } c = \max(\sigma)
\]
By using the standard deviation, \( \sigma_{p_k} \), NeuRN captures the contrast of local features, leading to representations that can generalize better across domains. Unlike Local Response Normalization (LRN) \cite{11}, which focuses on channel-wise activations using ratios of activations to neighbouring sums, and Local Contrast Normalization (LCN) \cite{Jarrett2009WhatIT}, which enhances individual pixel contrast but may miss the larger structural context, NeuRN considers spatial information of patches and maintains structural details throughout the image. This holistic, context-aware approach accommodates both local and global patterns, making NeuRN robust across varied visual settings. By aligning closely with neurobiological insights—specifically how neurons in the visual cortex encode contrasting features—NeuRN provides a more biologically plausible model for response normalization. In our analysis, we incorporate NeuRN primarily as a preprocessing layer.

\subsection{Needleman-Wunsch (N-W) algorithm}
The Needleman-Wunsch (N-W) algorithm \cite{42}, a foundational method in bioinformatics designed for sequence alignment, is adapted in our study to assess this structural similarity between different neural architectures by aligning layers of neural network models. For each pair of models, we define a scoring matrix \( M \) of dimensions \( A \times B \), where \( A \) and \( B \) are the number of layers in each respective neural network. A scoring system is essential for the alignment process, which comprises match/mismatch scores and a dissimilarity penalty \( d \) for introducing gaps during the alignment.
\begin{equation}
M_{i,j} =
\begin{cases}
0 & \text{if } i = 0 \text{ or } j = 0 \\
M_{i-1, j-1} + s(A_i, B_j) & \text{diagonal score} \\
M_{i-1, j} + d & \text{vertical} \\
&  \text{dissimilarity penalty} \\
M_{i, j-1} + d & \text{horizontal} \\
& \text{dissimilarity penalty} \\
\end{cases}
\end{equation}
\begin{itemize}
\item \( s(A_i, B_j) \) is the score when layer \( A_i \) is aligned with layer \( B_j \), based on their similarity. For example, if \( Conv_i \) represents a convolution layer in VGG19 and \( Conv_j \) represents a convolutional layer in VGG16, \( s(Conv_i, Conv_j) \) = 2 where 2 is the match score assigned as an input parameter since both layers represent 2D Convolution.
\item \( d \) is the dissimilarity penalty introduced when layers from one DNN are not aligned with those from another. For example, if \( Conv_i \) represents a convolution layer in VGG19 and \( ReLU_j \) represents a ReLU activation in VGG16, \( s(Conv_i, ReLU_j) \) = d = -2 where -2 is the mis-match penalty.
\end{itemize}
To compare DNNs, we encode each layer by assigning a unique identification. Each model is treated as a string encoding its architectural features identified by the type of layer. The recurrence relations used in the N-W algorithm are:
\begin{enumerate}
\item If the score in the cell \( M(i-1, j-1) \) and the score from matching the ith character of the first sequence with the jth character of the second sequence agree, a match\_score is awarded; otherwise, a penalty is applied.
\item The score in the cell \( M(i-1, j) \) is added to a penalty for inserting a gap in the second sequence.
\item The score in the cell \( M(i, j-1) \) is added to a penalty for inserting a gap in the first sequence.
\end{enumerate}
Based on this method, we shortlist 12 models from our initial selection of 44 DNNs (the resulting confusion matrix can be seen in {{\textbf{Figure \ref{fig:4}}}}). These models are: VGG19 \cite{13}, EfficientNetB0 \cite{32}, DenseNet121 \cite{huang2018densely}, ShuffleNet \cite{34}, Xception \cite{15}, NASNetMobile \cite{zoph2018learning}, ResNet50 \cite{resnet50}, ResNet50V2 \cite{resnet50}, InceptionV3 \cite{szegedy2015rethinking}, MobileNet \cite{howard2017mobilenets}, MobileNetV2 \cite{37} and ViT \cite{33}.

\begin{figure*}
    \centering
    \includegraphics[scale=0.7]{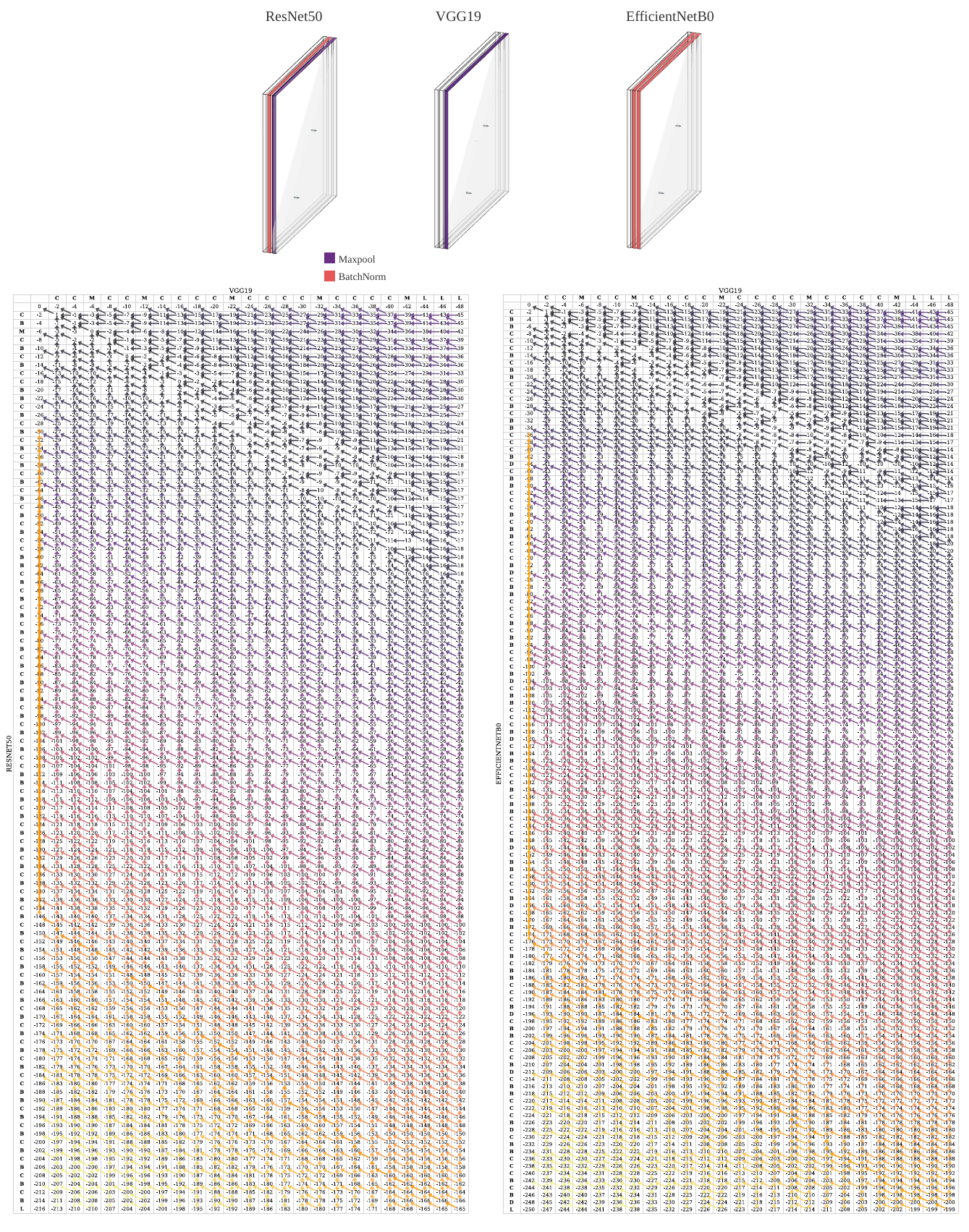}
    \caption{The dynamic matrix is constructed by filling in the cells iteratively using a set of recurrence relations between layers (denoted by characters). 
    In evaluating each cell we make a choice by selecting the maximum of three possibilities - a match, a dissimilarity or a gap. Hence we determine the value of each (uninitialized) cell in the matrix by the cell to its left, above, and its diagonally left. A match and a dissimilarity are both represented as traveling in the diagonal direction as shown by the pointers. Finally, we traceback through the choices in the matrix from the bottom-right cell to the top-left cell which gives us the similarity index between two models. Left Matrix shows how similarity index between ResNet50 and VGG19 was calculated and right matrix shows a similar process for VGG19 and EfficientNetB0.}
    \label{fig:dm}
\end{figure*}

\begin{figure*}
    \centering
    \includegraphics[scale=0.6]{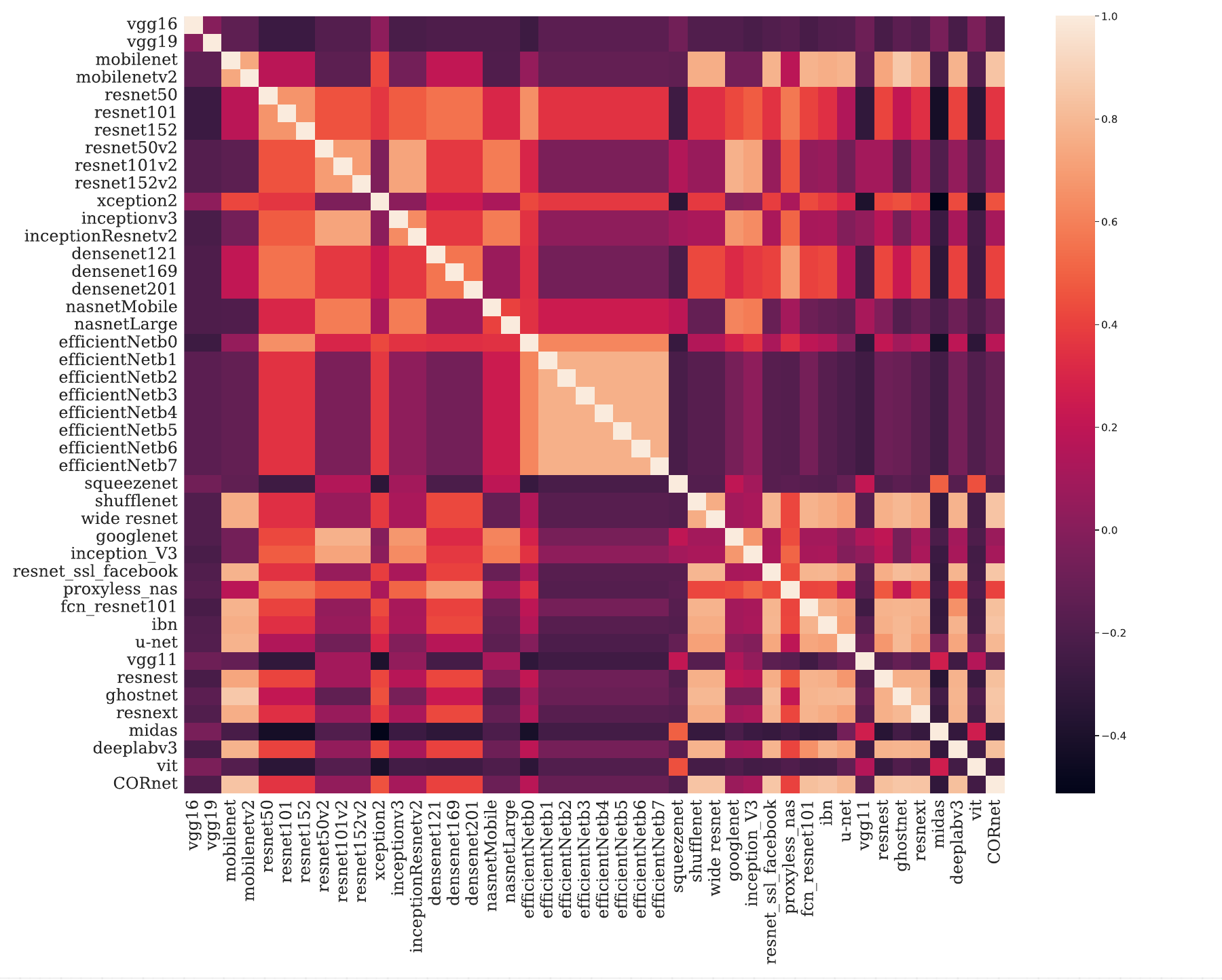}
    \caption{Each cell in this matrix represents the similarity index computed by number of common combinations across the entire model space. For instance, VGG19 contains 2xConv2D followed by a maxpool layer, any other models with this combination of layers in order would rank higher in similarity as compared to the ones that lack this combination.}
    \label{fig:pattern}
\end{figure*}

To further explain {{\textbf{Figure \ref{fig:4}}}}, we encode each layer by assigning a unique identification to every type of layer in a given deep neural network, \textit{Conv2D, BatchNormalization, Maxpool, and Dropout} are few examples of such layers. Diving further, {{\textbf{Figure \ref{fig:dm}}}} demonstrates another example of architectural comparison where the top section shows a set of layers from ResNet50, VGG19 and EfficientNetB0 highlighting dissimilarity of layers in the corresponding subsections. We treat each model as a string which encodes the architectural aspects of the model while also preserving its layer's functionality. Afterwards, we apply the Needleman-Wunsch algorithm ({{\textbf{Figure \ref{fig:dm}}}}) with a match\_score of 4, vertical and horizontal dissimilarity of -1 and build a matrix from left to right and top to bottom, using a set of recurrence relations to compute the score in each cell. For a matrix $M$ of size $l$ x $m$ where $l$ and $m$ are the lengths of strings that encode model architectures. The recurrence relations used in the Needleman-Wunsch algorithm are as follows:
\begin{enumerate}
  \item If the score in the cell $M(i-1, j-1)$ and the score resulting from matching the i-th character of the first sequence with the j-th character of the second sequence are in agreement, a match\_score is awarded. If not, a penalty is applied.
  \item The score in the cell $M(i-1, j)$ is added to a penalty for inserting a gap in the second sequence (indicating vertical dissimilarity).
  \item The score in the cell $M(i, j-1)$ is added to a penalty for inserting a gap in the first sequence (indicating horizontal dissimilarity).
\end{enumerate}
\vspace{0.1cm}
\noindent In order to further investigate about what type of layers are grouped together in the artificial neural networks, we also performed a pattern-based approach where we group together the existing layers in the model space and use these layer combinations to compare architectures in the model space. {{\textbf{Figure \ref{fig:scatter}}}} shows top 100 common combinations amongst the models in the model space showing there is a lot of alignment between the groups of DNN architectures. For every pair of models, we identify the common patterns and assign them a weighted score  based on the number of these intersecting combinations between the given two models. Here, the minimum length of a combination is 2 and the maximum length is $n-1$, where $n$ is the minimum number of layers present in either model. In addition, {{\textbf{Figure \ref{fig:pattern}}}} shows these results in a qualitative form. The resulting clusters of model similarity match our analysis in {{\textbf{Figure \ref{fig:4}}}} which shows how the pattern-based and layer-based approaches converge to the same results.

\begin{figure*}
    \centering
    \includegraphics[scale=0.035]{supp_figures/scatterplot_patterns.pdf}
    \caption{Each x-label in the plot represents a layer combination and each marker represents the DNN  which contains the corresponding combination of layers }
    \label{fig:scatter}
\end{figure*}

\subsection{Evaluation Datasets}
We evaluate domain generalization using MNIST \cite{1}, SVHN \cite{2}, USPS \cite{3}, and MNIST-M \cite{4}, each presenting unique challenges. MNIST offers 70,000 grayscale images of handwritten digits. SVHN provides over 600,000 images of house numbers in real-world contexts, adding complexity to digit recognition. USPS contains 9,298 16×16 grayscale images from scanned mail. MNIST-M merges MNIST digits with coloured backgrounds from BSDS500, introducing variations in colour and texture. These datasets test models’ ability to generalize from clean handwritten digits to numbers in varied, naturalistic settings, serving as benchmarks for vision models in diverse environments.

\section{Experiments \& Results}
\input{tables/table}

We shortlist a representative subset of DNN architectures—including CNNs, ViT, and NAS-derived models—to test NeuRN. By integrating our proposed layer into these networks, we aim to enhance their domain generalization capabilities. After embedding NeuRN, we conduct comprehensive benchmarking to evaluate its impact on performance across the selected models.

\subsection{Similarity across DNN architectures}
To select a representative sample that captures the architectural uniqueness among the vast variety of DNN architectures, we meaningfully sample from an expansive pool of image classification models while ensuring comprehensive representation. We adapt an architectural analysis approach focusing on the sequential arrangement and types of layers within DNNs. Inspired by \cite{20}, which employs spectral clustering of neuron groups across networks (e.g., \textit{Conv2D} clusters), we extend this concept from individual neurons to entire layers. Utilizing the N-W algorithm \cite{42}, we compute similarity scores between DNNs, forming architectural clusters within our sample space. By aligning DNN architectures as sequences using the N-W algorithm, we ascertain global similarities between paired models through recurrent relations between layers to compute a similarity index. This computation is performed for each pair among our collection of 44 DNN architectures, as depicted in \textbf{Figure \ref{fig:4}}. The results show that variants of similar architectures, such as VGG (\textbf{Figure \ref{fig:4}}), naturally group together in high-similarity clusters due to identical repeating layers. Additionally, we observe overlaps between variants of distinct models within the similarity space, likely due to recurrent layer combinations across these architectures.

\subsection{NeuRN-derived performance boost for domain generalization in DNNs}
Using digit datasets, we test NeuRN’s domain-agnostic feature encoding capabilities and evaluate its impact on domain generalization. Results are summarized in \textbf{Table 1}. DNNs fine-tuned on a single source domain and tested on various target domains show significant improvement in cross-domain evaluation when equipped with NeuRN. This is particularly evident in transitions from MM$\rightarrow$S, M$\rightarrow$S, M$\rightarrow$MM, and U$\rightarrow$S, highlighting NeuRN’s strong domain bridging capability.
\\\\
\noindent\textbf{VGG19} outperforms in 9 out of 12 domain transfer tasks, registering substantial performance enhancements in transitions such as M $\rightarrow$ MM (from 39.7\% to \textbf{62.6\%}), U $\rightarrow$ MM (from 19.9\% to \textbf{30.5\%}), and MM $\rightarrow$ S (from 22.5\% to \textbf{33.6\%}). It achieves a formidable accuracy of \textbf{96.2\%} in MM $\rightarrow$ M, nearing the competitive benchmark. \textbf{EfficientNetB0} demonstrates a performance boost in 7 out of 12 domain transfer tasks. A notable leap in performance is seen in MM $\rightarrow$ U, where accuracy improves from 17.8\% to \textbf{51.3\%}. \textbf{DenseNet121} achieves performance enhancement in 8 out of 12 tasks, DenseNet121 shows significant performance increases, particularly in M $\rightarrow$ MM (from 38.4\% to \textbf{50.8\%}) and S $\rightarrow$ M (from 34.1\% to \textbf{49.1\%}). \textbf{ShuffleNet} improves in 8 out of 12 tasks, demonstrates a robust performance increase in MM $\rightarrow$ M (from 38.7\% to \textbf{84.8\%}) and M $\rightarrow$ MM (from 14.1\% to \textbf{71.1\%}). Notable improvements are also observed in U $\rightarrow$ S (from 12.2\% to \textbf{36.2\%}) and MM $\rightarrow$ S (from 18.3\% to \textbf{50.4\%}). \textbf{Xception} excels in 9 out of 12 tasks with significant gains, particularly in U $\rightarrow$ M, elevating accuracy from 20.9\% to \textbf{41.3\%}. \textbf{NASNetMobile} exhibits enhanced performance in 9 out of 12 tasks, with remarkable boosts in M $\rightarrow$ U, M $\rightarrow$ MM, and a competitive accuracy of \textbf{95.6\%} in MM $\rightarrow$ M. \textbf{ResNet50} improves in 7 out of 12 tasks, with a significant increase in M $\rightarrow$ MM from 48.3\% to \textbf{60.1\%}. \textbf{ResNet50V2} excels in 9 out of 12 tasks, notably in U $\rightarrow$ M (from 11.1\% to \textbf{28.4\%}), and MM $\rightarrow$ S (from 17.2\% to \textbf{31.3\%}). \textbf{InceptionV3} achieves a performance increase in 6 out of 12 tasks, significantly increases performance in U $\rightarrow$ S (from 34.2\% to \textbf{50.8\%}) and in MM $\rightarrow$ S (from 15.0\% to \textbf{34.0\%}). \textbf{MobileNet} achieves a performance boost in 7 out of 12 tasks, significantly boosts performance in MM $\rightarrow$ S, jumping from 23.4\% to \textbf{50.7\%}. \textbf{MobileNetV2} improves in 9 out of 12 tasks, attaining competitive benchmarking performance in MM$ \rightarrow$ M with \textbf{96.2\%} accuracy and in MM $\rightarrow$ U, escalating from 44.9\% to \textbf{88.5\%}. \textbf{ViT} achieves a performance boost in 6 out of 12 tasks, with a pronounced increase in the M $\rightarrow$ MM task from 32.8\% to \textbf{44.7\%} accuracy.

\subsection{NeuRN-derived performance boost for domain generalization in NAS-derived architectures} Building on promising results with DNNs, we explore NeuRN’s domain generalization capabilities in complex architectures like NAS methods. NAS automatically designs neural network architectures tailored to specific tasks by searching over a predefined space guided by performance metrics. We examine two prominent NAS strategies: Single Path One-Shot (SPOS) \cite{30} and Autoformer \cite{31}. SPOS streamlines CNN architecture search by focusing on single-path evaluations within a superstructure. Autoformer innovates on ViT architecture, automating configuration through a one-shot search within a super network, allowing derived models to inherit finely tuned weights. As seen in \textbf{Table 1}, NeuRN’s integration with SPOS leads to remarkable improvements in 10 out of 12 domain transfer tasks. For example, U$\rightarrow$M accuracy increased to \textbf{83.4\%}, and MM$\rightarrow$M reached \textbf{98.4\%}. MM$\rightarrow$U achieved \textbf{88.4\%} accuracy, and MM$\rightarrow$S improved from 25.6\% to \textbf{69.0\%}. These enhancements demonstrate NeuRN’s ability to facilitate high-performance domain generalization, overcoming limitations in NAS models that often struggle with domain shifts. Similarly, Autoformer architectures benefit from NeuRN’s integration, showing significant gains in 7 out of 12 tasks. It achieved \textbf{93.8\%} accuracy for U$\rightarrow$M and \textbf{90.2\%} for MM$\rightarrow$U, with MM$\rightarrow$S increasing from 24.6\% to \textbf{42.1\%}. These results highlight NeuRN’s potential to enhance domain generalization in architectures relying on self-attention mechanisms central to transformers.
\\\\
\noindent Empirical evidence shows that NeuRN substantially enhances DNN performance across various domain transfers, though not in all cases—highlighting the need for future investigation into model behavior post-NeuRN integration. The observed performance boost suggests that NeuRN enables models to generate domain-agnostic feature representations, reflecting adaptive processes in biological excitatory neurons. Similar to the visual cortex’s ability to encode high-frequency features \cite{24}, our experiments reveal that integrating NeuRN allows models to effectively translate learned features between domains, adeptly handling increased complexity or distinctness in the target domain. This robust adaptability is especially prominent in the MM$\rightarrow$S task, where NeuRN significantly enhances performance. Such adaptability underscores the potential of NeuRN-enhanced models to serve as a foundation for more versatile and robust AI systems proficient in various visual domains.

\section{Conclusion}
Our work introduces two key contributions: (i) a unique framework for assessing the similarity between various DNN architectures and (ii) a brain-inspired technique for domain generalization. Our proposed framework serves the purpose of identifying similar DNN architectures. We use the N-W algorithm to provide a quantitative, architecture-based comparative analysis. This approach allows for an informed selection of architectures to form a representative set of existing DNNs for an efficient yet comprehensive analysis. We introduce NeuRN as a pre-processing layer to enhance DNNs' ability to learn domain-agnostic representations which would aid generalization across domains. We fine-tune DNNs and evaluate domain generalization performance across unseen domains. We repeat this process for the same DNNs integrated with NeuRN as a pre-processing layer. The DNNs include CNNs, ViT, and NAS-derived models. Results show that NeuRN improves the domain generalization capabilities of DNNs in image classification tasks, boosting performance across multiple domain transfer tasks. 

However, extending NeuRN experimentation across additional high-resolution domain generalization datasets requires additional resources and forms a limitation of our study. Our work carries significant implications, envisioning a future where domain generalization is a natural feature integrated into neural network design, rather than a challenge to overcome.

\input{aaai25.bbl}
\clearpage
\renewcommand{\figurename}{Supplementary Figure}
\renewcommand{\tablename}{Supplementary Table}

\setcounter{figure}{0}
\setcounter{table}{0}

\renewcommand{\thefigure}{S\arabic{figure}}
\renewcommand{\thetable}{S\arabic{table}}

\section{Appendix}
\section{DNNs’ architectural and functional similarity}
In our comprehensive main analysis, we observe notable correlations between the architectural design and functional performance of various DNNs. To further explain these correlations, we conduct a row vector analysis of the performance data presented in {{\textbf{Table 1}}}. This analysis utilizes cosine similarity scores to quantify the functional similarity between different models based on their performance accuracies for all domain-shift tasks.

The integration of NeuRN into various DNNs has a profound impact on their functional similarity, as measured by the performance data analysis with and without integration of NeuRN. This effect is noticeable in {{\textbf{Supplementary Figure S1}}} where we show the difference in cosine similarity scores (NeuRN-integrated models $-$ non-NeuRN models). Prior to the integration of NeuRN, the average functional similarity score among the models stood at approximately \textbf{0.7}. This score reflects the overall degree of similarity in performance across different models without any enhancement. However, with the integration of NeuRN, this average functional similarity score notably increases to \textbf{0.8}. This increase highlights NeuRN's capability not just in elevating individual model performance but also in bringing models closer in terms of functional behavior. The increased functional similarity with NeuRN integration suggests that our novel technique aids in harmonizing the performance characteristics of diverse models. As seen in our performance data ({{\textbf{Table 1}}}), NeuRN consistently enhances model performance across a variety of domain transfer tasks. More importantly, this enhancement does not disproportionately favor one model over another. Instead, it uniformly elevates performance levels, thereby reducing the performance disparity between different architectural designs. This finding is significant as it indicates that NeuRN not only boosts the overall performance metrics of individual models but also promotes a more uniform functional behavior across various architectures. Consequently, the integration of NeuRN emerges as a powerful tool in the domain of DNNs, not only for performance enhancement but also for fostering greater functional coherence among diverse neural network architectures. This phenomenon aligns with the earlier observed correlations between architectural design and functional performance ({\textbf{Table 1}}). 

\section{Training details}
\subsection{Neural Architecture Search (NAS) setup}
\begin{figure}
    \centering
    \begin{minipage}{\textwidth}
    \centering
    \includegraphics[scale=0.7]{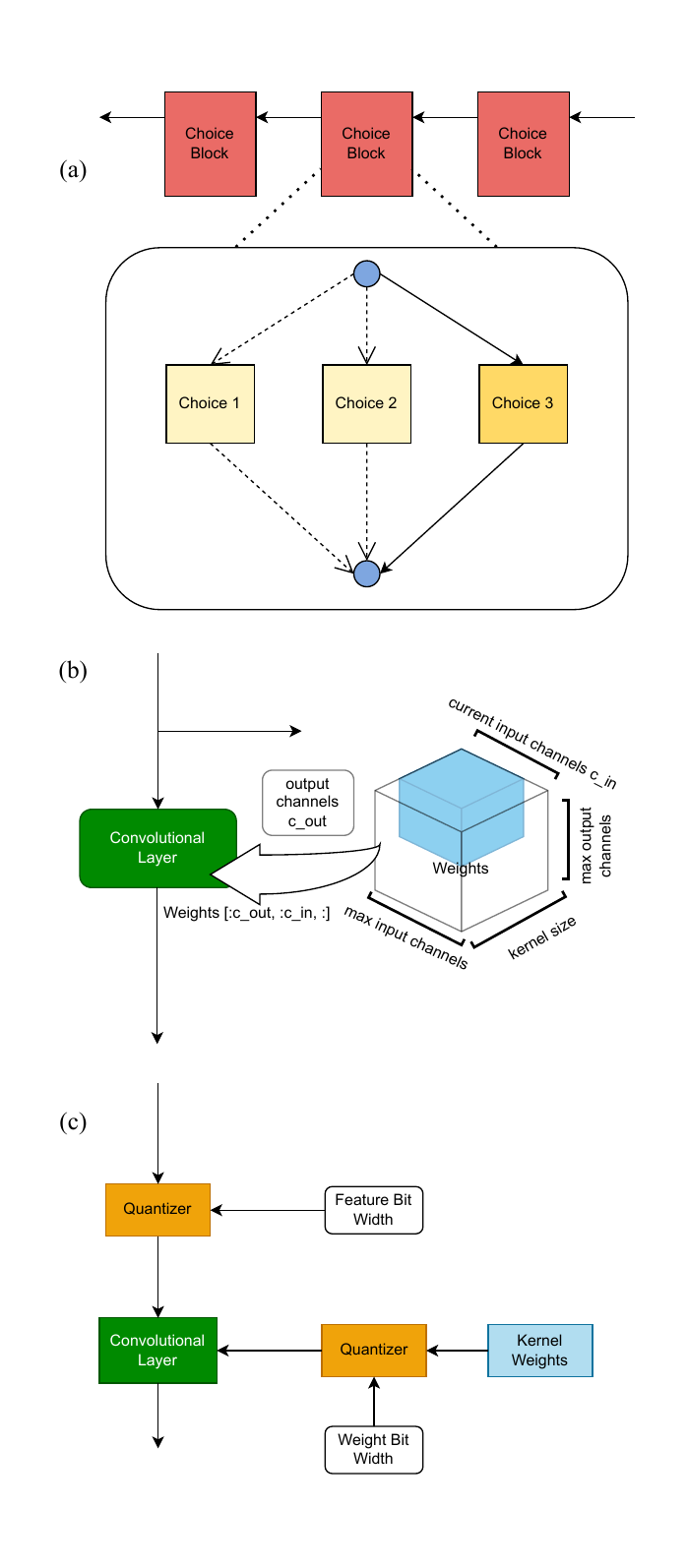}
    \end{minipage}
    \begin{minipage}{\textwidth}
        \vspace{10mm}
        \centering
        \caption{Overview of the Single Path One-Shot (SPOS) supernet architecture and the novel choice block designs. (a) At the top, the general layout of a choice block is depicted, where one architecture choice is invoked at a time to form a stochastic architecture. (b) The middle figure showcases the proposed choice block for Channel Number Search, which utilizes a weight-sharing approach and facilitates faster convergence of the supernet. The block preallocates a weight tensor with a maximum number of channels and randomly selects the required channels for convolution during training. (c) The bottom figure represents the mixed-precision quantization search block that allows for bit width search of weights and feature representations. This block also integrates with the channel search space to enhance the efficiency of the search process. Specific bit widths are randomly sampled during supernet training and determined during the search step.}
        \label{fig:spos}
        \end{minipage}
\end{figure}

\begin{figure}
    \centering
    \begin{minipage}{\textwidth}
        \centering
        \hspace{-20mm}
        \includegraphics[scale=0.75]{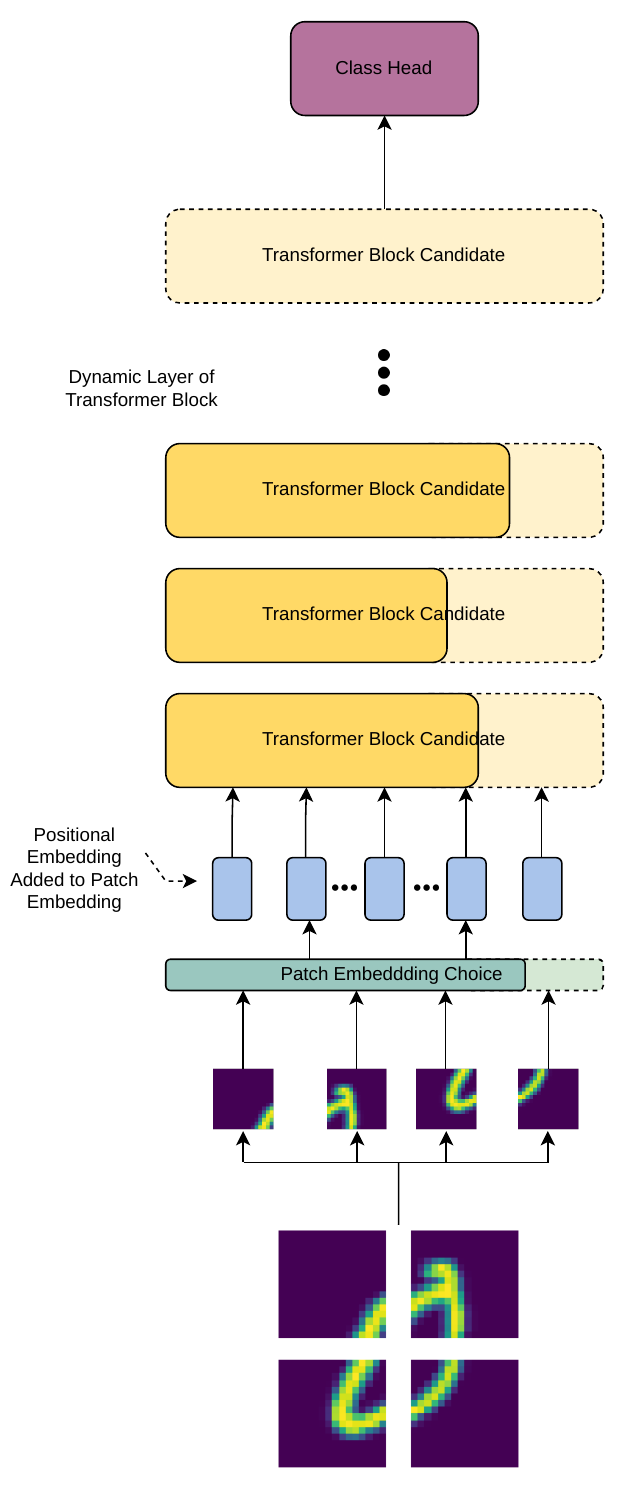}
    \end{minipage}
    \begin{minipage}{\textwidth}
        \centering
        \caption{MNIST dataset was used to fine-tune NAS Supernet of Autoformer. The Supernet-S pre-trained on ImageNet,  Image size of 32x32 was broken down into 4 patches of 16 x 16 which are then used to find positional embeddings. Positional embeddings are combined with patch embedding to train the supernet. Once Supernet is finetuned an evolution search is performed to obtain the best performing model. AutoFormer employs a "weight entanglement" approach. During training, weights from overlapping candidate blocks are simultaneously updated, represented by $w^{(i)}{j} \subseteq w^{(i)}{k}\text{or }w^{(i)}{k} \subseteq w^{(i)}{j}$, where weights from block j and block k in the layer i intersect. A subnet is sampled from the search space, and its weights are updated while the rest are frozen.Post-SuperNet training, both approaches employ an evolutionary search to find the optimal architecture, seeking to maximize validation accuracy with minimal model size, as represented by $a^{*} = \arg\max_{a \in A} \text{ACCA}{\text{val}}(N(a, W{A}(a)))$  }
        \label{fig:auto}
    \end{minipage}
\end{figure}

Our study extends the application of the biologically-inspired layer to the novel architectures, optimized using Neural Architecture Search. For this purpose, we applied two strategies: Single Path One-Shot (SPOS) \cite{30} ({{\textbf{Supplementary Figure \ref{fig:spos}}}}) and AutoFormer \cite{31} ({{\textbf{Supplementary Figure \ref{fig:auto}}}}). The first stage of SuperNet training is common across both techniques, followed by an evolutionary search to generate optimized architectures. 

AutoFormer employs a "weight entanglement" approach. During training, weights from overlapping candidate blocks are simultaneously updated, represented by $w^{(i)}{j} \subseteq w^{(i)}{k}\text{or }w^{(i)}{k} \subseteq w^{(i)}{j}$, where weights from block $j$ and block $k$ in the layer $i$ intersect. A subnet is sampled from the search space, and its weights are updated while the rest of the weights are frozen.

On the contrary, SPOS decouples weight and architecture optimization. Weights are optimized via weight sharing, represented by $W_{A} = \arg\min_{W} \mathcal{L}_{\text{train}}(N(A,W))$. It is a hyperparameter-free optimization, as it randomly selects a sub-network of the SuperNet for evaluation during training. 
Post-SuperNet training, both approaches employ an evolutionary search to find the optimal architecture, seeking to maximize validation accuracy with minimal model size, as represented by $a^{*} = \arg\max_{a \in A} \text{ACCA}{\text{val}}(N(a, W{A}(a)))$.

\subsection{DNNs' setup}
Both existing and NAS-derived DNN models are employed in our experimental framework. Leveraging ImageNet pre-trained weights, we applied these models to the MNIST datasets for domain generalization tasks. Throughout the training phase, early stopping was consistently applied to prevent over-fitting and to optimize performance. In order to train the DNN models, a learning rate of 0.001, batch size of 256 and Adam optimizer was used. For SPOS, A learning rate of 0.5, batch size of 128 and SGD optimizer was used in training. Autoformer was used with a learning rate of 0.01, batch size of 24 and Adam optimizer was used to run training. Early stopping patience of 5 was used with validation accuracy as a performance metric. All experiments were performed using NVIDIA TITAN RTX GPU.

\begin{figure*}
    \centering
    \includegraphics[scale=0.6]{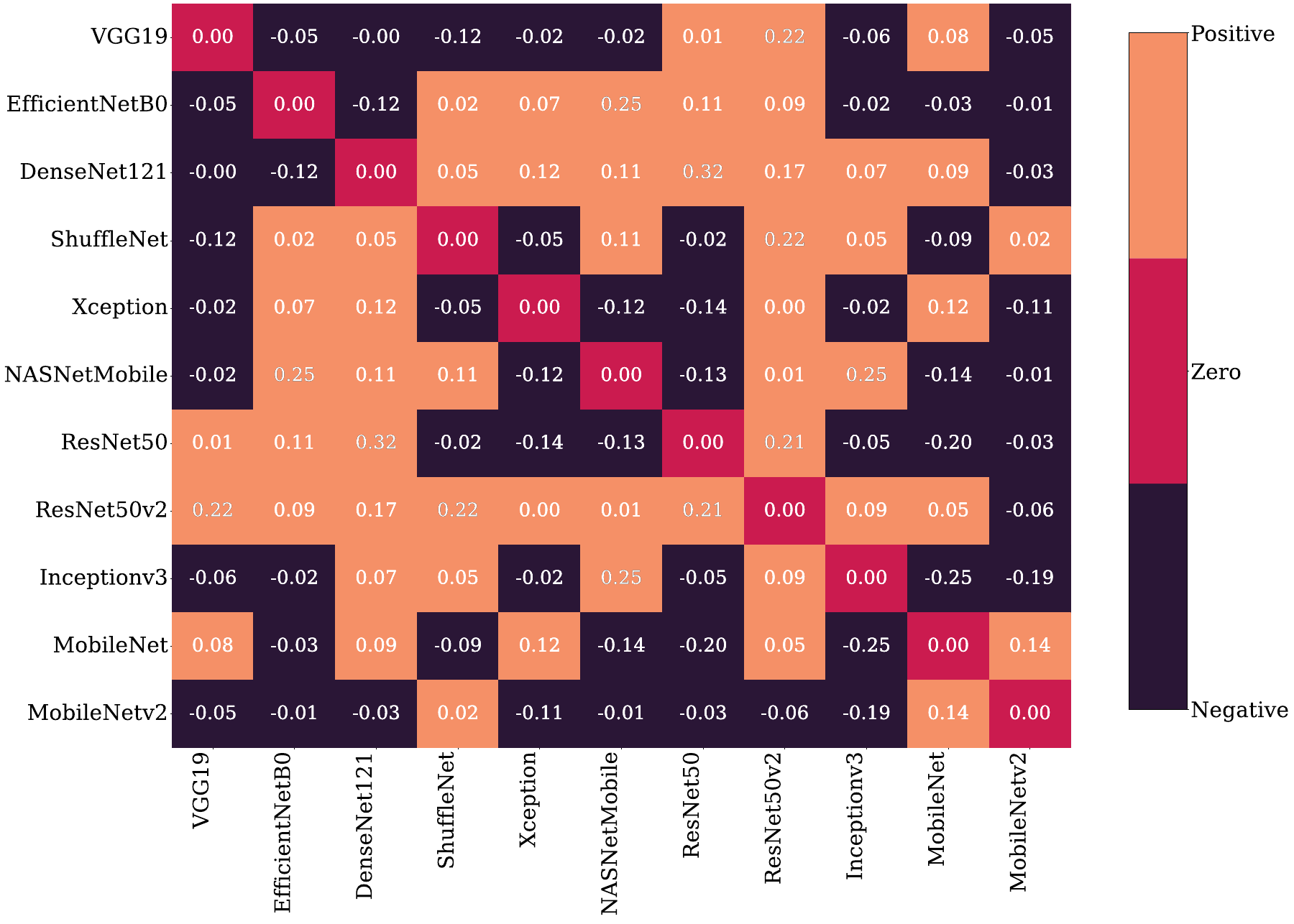}
    \caption{Based on {{\textbf{Table 1}}}, this matrix depicts the pairwise functional similarity of shortlisted deep neural network models resulting from the integration of NeuRN. The similarity is quantified using cosine similarity scores, derived from the models' performance on domain transfer tasks from MNIST datasets in {{\textbf{Table 1}}}. Positive cells represent increased functional similarity for NeuRN integrated models, Zero indicates no difference, while Negative indicates dissimilarity.}
    \label{fig:woutneurn}
\end{figure*}


\clearpage

\end{document}

%% file: tables/table.tex
\begin{table*}[htbp]
\centering
\scalebox{0.75}{
\label{tab:table}
\begin{tabular}{l|*{12}{c}}
\toprule
\textbf{Models} & \rotatebox{45}{\textbf{M$\rightarrow$U}} & \rotatebox{45}{\textbf{M$\rightarrow$S}} & \rotatebox{45}{\textbf{M$\rightarrow$MM}} & \rotatebox{45}{\textbf{U$\rightarrow$M}} & \rotatebox{45}{\textbf{U$\rightarrow$S}} & \rotatebox{45}{\textbf{U$\rightarrow$MM}} & \rotatebox{45}{\textbf{S$\rightarrow$M}} & \rotatebox{45}{\textbf{S$\rightarrow$U}} & \rotatebox{45}{\textbf{S$\rightarrow$MM}} & \rotatebox{45}{\textbf{MM$\rightarrow$M}} & \rotatebox{45}{\textbf{MM$\rightarrow$U}} & \rotatebox{45}{\textbf{MM$\rightarrow$S}} \\
\midrule

VGG19 & 
70.9 & 
13.7 & 
39.7 & 
66.2 & 
13.9 & 
19.9 & 
47.2 & 
32.5 & 
33.9 & 
94.3 & 
76.7 & 
22.5 \\

VGG19+NeuRN & 
\textbf{74.2} & 
\textbf{24.5} & 
\textbf{62.6} & 
48.4 & 
\textbf{20.4} & 
\textbf{30.5} & 
\textbf{51.3} & 
28.4 & 
\textbf{38.4} & 
\underline{\textbf{96.2}} & 
74.3 & 
\textbf{33.6} \\
\hline

EfficientNetB0 & 
8.6 & 
7.4 & 
18.4 & 
9.4 & 
8.0 & 
10.6 & 
34.6 & 
12.0 & 
18.6 & 
96.8 & 
17.8 & 
12.5 \\

EfficientNetB0+NeuRN & 
7.6 & 
6.8 & 
18.2 & 
\textbf{11.6} & 
\textbf{10.8} & 
\textbf{10.8} & 
24.2 & 
\textbf{14.9} & 
\textbf{21.1} & 
93.1 & 
\textbf{51.3} & 
\textbf{21.8} \\
\hline

DenseNet121 & 
74.3 & 
11.7 & 
38.4 & 
39.5 & 
11.4 & 
20.9 & 
34.1 & 
22.1 & 
31.5 & 
96.6 & 
73.3 & 
21.3 \\

DenseNet121+NeuRN & 
26.4 & 
\textbf{15.9} & 
\textbf{50.8} & 
\textbf{43.1} & 
\textbf{14.8} & 
\textbf{21.0} & 
\textbf{49.1} & 
20.0 & 
\textbf{32.6} & 
85.2 & 
54.0 & 
\textbf{26.3} \\
\hline

ShuffleNet & 
9.3 & 
19.0 & 
14.1 & 
44.9 & 
12.2 & 
13.0 & 
26.1 & 
31.1 & 
19.7 & 
38.7 & 
3.2 & 
18.3 \\
ShuffleNet+NeuRN & 
7.0 & 
\textbf{56.1} & 
\textbf{71.1} & 
22.6 & 
\textbf{36.2} & 
\textbf{27.2} & 
\textbf{32.1} & 
27.5 &
\textbf{28.9} & 
\underline{\textbf{84.8}} & 
\textbf{7.6} & 
\textbf{50.4} \\

\hline

Xception & 
62.4 & 
19.4 & 
56.0 & 
20.9 & 
11.9 & 
21.1 & 
42.7 & 
14.2 & 
34.2 & 
97.9 & 
77.8 & 
25.6 \\

Xception+NeuRN & 
61.9 & 
\textbf{20.9} & 
\textbf{59.7} & 
\textbf{41.3} & 
\textbf{15.9} & 
\textbf{25.8} & 
\textbf{48.8} & 
\textbf{16.4} & 
\textbf{37.7} & 
95.7 & 
31.8 & 
\textbf{33.1} \\

\hline

NASNetMobile & 
33.0 & 
12.7 & 
36.0 & 
24.8 & 
10.1 & 
21.4 & 
46.0 & 
23.8 & 
32.6 & 
90.0 & 
63.7 & 
20.3 \\

NASNetMobile+NeuRN & 
\textbf{45.4} & 
\textbf{21.1} & 
\textbf{47.5} & 
\textbf{25.5} & 
\textbf{10.4} & 
16.9 & 
\textbf{47.6} & 
16.5 & 
\textbf{35.1} & 
\underline{\textbf{95.6}} & 
58.3 & 
\textbf{31.1} \\
\hline

ResNet50 & 
24.3 & 
15.9 & 
48.3 & 
37.2 & 
7.1 & 
20.8 & 
34.4 & 
10.6 & 
34.7 & 
96.5 & 
71.5 & 
21.2 \\

ResNet50+NeuRN & 
\textbf{31.3} & 
\textbf{21.8} & 
\textbf{60.1} & 
11.3 & 
6.1 & 
11.3 & 
\textbf{44.3} & 
\textbf{19.3} & 
\textbf{35.3} & 
96.0 & 
59.0 & 
\textbf{30.9} \\
\hline

ResNet50v2 & 
78.7 & 
21.6 & 
57.8 & 
11.1 & 
8.6 & 
10.5 & 
48.0 & 
15.6 & 
33.7 & 
97.2 & 
78.1 & 
17.2 \\

ResNet50v2+NeuRN & 
31.4 & 
\textbf{23.1} & 
\textbf{59.6} & 
\textbf{28.4} & 
\textbf{14.1} & 
\textbf{20.7} & 
\textbf{50.5} & 
\textbf{18.8} & 
\textbf{35.7} & 
96.4 & 
55.7 & 
\textbf{31.3} \\
\hline

InceptionV3 & 
69.8 & 
16.4 & 
57.8 & 
34.2 & 
8.9 & 
24.9 & 
53.5 & 
28.0 & 
39.0 & 
97.2 & 
82.0 & 
15.0 \\

InceptionV3+NeuRN & 
52.6 & 
\textbf{22.6} & 
\textbf{62.2} & 
\textbf{50.8} & 
\textbf{14.4} & 
\textbf{31.3} & 
47.9 & 
19.7 & 
37.6 & 
96.1 & 
75.5 & 
\textbf{34.0} \\
\hline

MobileNet & 
17.7 & 
17.0 & 
53.6 & 
16.2 & 
6.9 & 
16.6 & 
51.1 & 
18.0 & 
36.7 & 
97.0 & 
69.1 & 
23.4 \\

MobileNet+NeuRN & 
\textbf{25.8} & 
\textbf{24.5} & 
\textbf{58.5} & 
\textbf{17.6} & 
\textbf{9.4} & 
14.2 & 
45.3 & 
\textbf{19.2} & 
35.8 & 
96.4 & 
34.2 & 
\textbf{50.7} \\

\hline

MobileNetv2 & 
8.8 & 
16.1 & 
45.9 & 
32.6 & 
9.3 & 
17.8 & 
45.9 & 
17.7 & 
30.8 & 
94.2 & 
44.9 & 
20.8 \\

MobileNetv2+NeuRN & 
\textbf{20.9} & 
\textbf{16.3} & 
45.0 & 
\textbf{34.2} & 
\textbf{10.6} & 
17.5 & 
\textbf{48.9} & 
15.0 & 
\textbf{36.2} & 
\underline{\textbf{96.2}} & 
\underline{\textbf{88.5}} & 
\textbf{33.9} \\
\hline

ViT & 
69.4 & 
13.1 & 
32.8 & 
26.2 & 
9.4 & 
15.6 & 
45.2 & 
36.5 & 
30.6 & 
95.5 & 
64.4 & 
21.8 \\

ViT+NeuRN & 
61.9 & 
\textbf{16.2} & 
\textbf{44.7} & 
\textbf{36.6} & 
\textbf{14.1} & 
\textbf{25.3} &
33.2 & 
25.5 & 
26.5 & 
87.8 & 
58.9 & 
\textbf{23.5} \\
\hline

SPOS (NAS) & 17.8
& 93.2 
& 20.5
& 79.6
& 16.6
& 21.1
& 70.0
& 66.2 
& 51.5
& 95.9
& 81.8
& 25.6 \\

SPOS+NeuRN (NAS) & 
\textbf{38.3}
& 90.8
& \textbf{55.0}
& \underline{\textbf{83.4}}
& \textbf{34.7}
& \textbf{46.5}
& \textbf{75.5}
& 50.9
& \textbf{59.5}
& \underline{\textbf{98.4}}
& \underline{\textbf{88.4}}
& \textbf{69.0} \\
\hline

Autoformer (NAS) & 
95.2 & 
26.2 & 
79.8 & 
92.9 & 
30.7 & 
68.4 & 
56.5 & 
47.8 & 
45.3 & 
99.1 & 
83.6 & 
24.6 \\

Autoformer+NeuRN (NAS) & 
91.0 & 
\textbf{38.0} & 
73.5 & 
\underline{\textbf{93.8}} & 
\textbf{36.9} & 
66.1 & 
\textbf{60.5} & 
41.2 & 
\textbf{48.9} & 
98.5 & 
\underline{\textbf{90.2}} & 
\textbf{42.1} \\

\bottomrule
\end{tabular}
}
\caption{The table compares shortlisted fine-tuned CNNs, ViT and NAS-derived (NAS) DNNs performance on domain transfer tasks. Rows list models, and columns specify source $\rightarrow$ target domains: MNIST (M), SVHN (S), USPS (U), and MNIST-M (MM). Bold entries show performance improvements with NeuRN. Bold and underlined entries show performance improvements which are also close to benchmark results.}

\end{table*}